\documentclass[10pt,twocolumn,letterpaper]{article}

\usepackage{cvpr}
\usepackage{times}
\usepackage{epsfig}
\usepackage{graphicx}
\usepackage{amsmath}
\usepackage{amssymb}
\usepackage{multirow}
\usepackage{pifont}
\usepackage{subfigure}

\usepackage[breaklinks=true,bookmarks=false]{hyperref}

\cvprfinalcopy 

\def\cvprPaperID{****} 

\begin{document}

\title{BoostGAN for Occlusive Profile Face Frontalization and Recognition}

 \author {Qingyan Duan\(^1\), Lei Zhang\(^1\)*\\
              { \(^1\)School of Microelectronics and Communication Engineering, Chongqing University, Chongqing, China}   \\
                  \tt\small {qyduan@cqu.edu.cn, leizhang@cqu.edu.cn,}
                  }

\maketitle

\begin{abstract}
   There are many facts affecting human face recognition, such as pose, occlusion, illumination,
   age, etc. First and foremost are large pose and occlusion problems, which can even result in more than 10\% performance degradation. Pose-invariant feature representation and face frontalization with generative adversarial networks (GAN) have been widely used to solve the pose problem. However, the synthesis and recognition of occlusive
   but profile faces is still an uninvestigated problem. To address this issue, in this paper, we aim to contribute an effective solution on how to recognize occlusive but profile faces, even with facial keypoint region (e.g. eyes, nose, etc.) corrupted. Specifically, we propose a boosting Generative Adversarial
   Network (\textit{\textbf{BoostGAN}}) for de-occlusion, frontalization, and recognition of faces. Upon the assumption that facial occlusion is partial and incomplete, multiple patch occluded images are fed as inputs for knowledge boosting, such as identity and texture information. A new aggregation structure composed of a deep GAN for coarse face synthesis and a shallow boosting net for fine face generation is further designed. Exhaustive experiments demonstrate that the proposed approach not only presents clear
   perceptual photo-realistic results but also shows state-of-the-art recognition performance for occlusive but profile faces.
\end{abstract}

\section{Introduction}
In recent years, face recognition has received a
great progress, enabled by deep learning techniques~\cite{Sun2014Deep}. However, there are still many problems remaining to be unresolved
satisfactorily. First and foremost is the large pose variation, a bottleneck in face recognition. Existing methods addressing pose variations can be divided into two main categories. One category aims to obtain pose-invariant features by hand-crafting
or deep learning on multi-view facial images~\cite{Kan2016Multi,Schroff2015FaceNet,
Liu2017SphereFace,Wang2018CosFace}. The other one, inspired by generative adversarial network (GAN) techniques, employs synthesis approaches to generate frontal view images~\cite{Hassner2014Effective, Sagonas2015Robust, Luan2017Disentangled,Huang2017Beyond,Hu2018pose}. For the latter, the generator in GANs
generally follows an encoder-decoder convolutional neural network architecture, where the encoder extracts identity preserved features and the decoder outputs realistic faces in target pose. Not only pose variation, but illumination difference was also considered for pose-invariant facial feature representation, synthesis, and recognition~\cite{Luan2017Disentangled}.

\begin{figure}
\centering
\subfigure[Profile]{
\includegraphics[width=1.5cm]{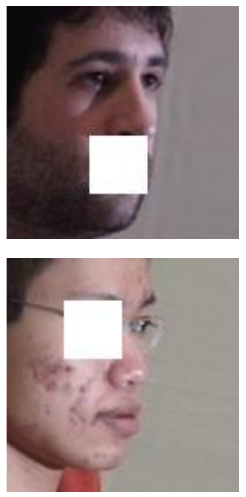}}
\subfigure[\textbf{Ours}]{
\includegraphics[width=1.5cm]{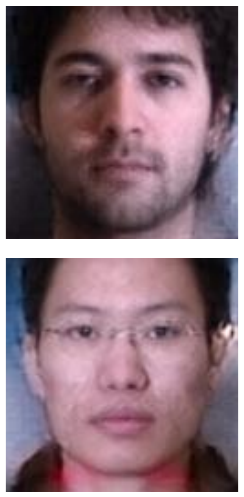}}
\subfigure[\cite{Luan2017Disentangled}]{
\includegraphics[width=1.5cm]{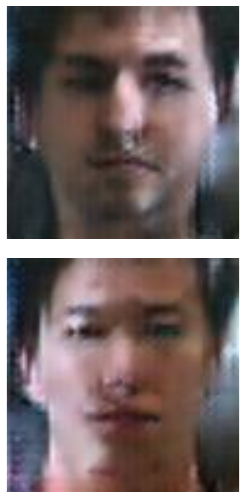}}
\subfigure[\cite{Huang2017Beyond}]{
\includegraphics[width=1.5cm]{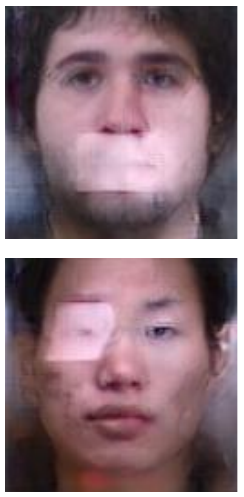}}
\subfigure[GT]{
\includegraphics[width=1.5cm]{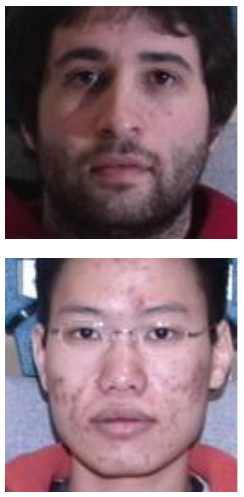}}
\caption{Synthesis results by testing the existing models on occluded faces. The poses of the $1^{st}$ row and the $2^{nd}$ row are $45^{\circ}$ and $60^{\circ}$, respectively. GT denotes the ground truth frontal images.}
\label{fig:f1}
\end{figure}

\begin{figure*}[t]
\centering
   \includegraphics[width=0.7\linewidth]{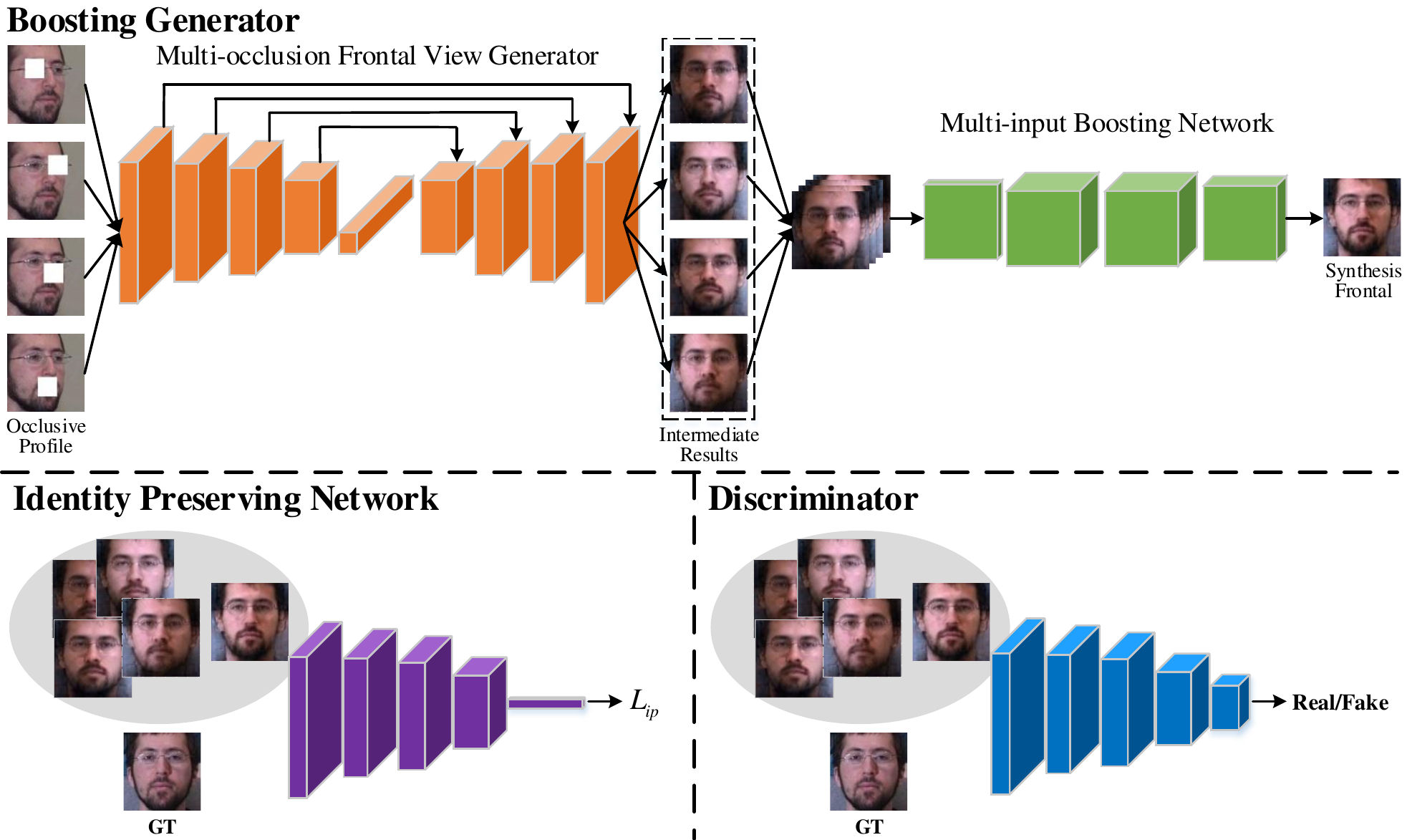}
   \caption{The framework of BoostGAN, in an end-to-end and coarse-to-fine architecture. Two parts are included: multi-occlusion frontal view generator and multi-input boosting network. For the former, the coarse de-occlusion and slight identity preserving images are generated. For the latter, the photo-realistic, clean, and frontal faces are achieved by ensemble complementary information from multiple inputs in a boosting way.}
\label{fig:f2}
\end{figure*}
In fact, except for pose variation, occlusion also seriously affects the performance of face recognition.
In order to fill the `hole' in faces, image completion is generally considered. Most traditional low-level cues based approaches synthesize the contents
by searching patches from the known region of the same image~\cite{Bertalmio2003Simultaneous,
Levin2003Learning,Wexler2007Space}. Recently, GAN has been used for image completion and achieved a success~\cite{Li2017Generative,Ishikawa2017Globally,Chen2018High}. However, these methods usually only generate the pixels of the corrupted region, while keeping other pixels of the clean region unchanged. It means that, these approaches are more appropriate for close-set image completion, because the
corrupted part in test images can not find the matched clean
image in training stage for open-set image completion. Therefore, the filled parts of testing images by these methods are imprecise and lack of identity discrimination. That is, if an occlusive
facial image is excluded in the training set, the repaired face cannot preserve the original
identity, which is not conducive to face recognition. Although Zhao et al.~\cite{Zhao2018Identity} introduced
an identity preservation loss, it is still not enough for
solving open-set image completion problem faultlessly.

To address this issue, in this paper, we aim to answer \textit{how to recognize faces if large pose variation and occlusion exist simultaneously}? This is a general problem in face recognition community. Briefly, our solution is the proposed BoostGAN model. To our knowledge, this is the first work for both de-occlusion and frontalization of faces by using GAN based variants with regard to face recognition. The previous face frontalization methods usually synthesize frontal view faces from clean profile
image. However, once the keypoint region is occluded or corrupted, the synthesis effect of these previous approaches becomes very poor (as is shown in Figure~\ref{fig:f1}). Additionally, the previous image completion methods were only used for occlusion part restoration of a \textit{near-frontal} view face, but not face frontalization. Although it can be used for filling the `hole' (de-occlusion) with GAN based image completion approaches, the generated faces can not well preserve identity and texture information. Consequently, the performance of occlusive profile face recognition is seriously flawed, especially when the key region, such as eyes,
mouth, is sheltered, as shown in Figure~\ref{fig:f1}.

Specifically, we propose an end-to-end BoostGAN model to solve the face frontalization when both occlusions and pose variations exist simultaneously. The identity preservation of the generated faces is also qualified for occlusive but profile face recognition. The architecture of BoostGAN is shown in Figure~\ref{fig:f2}. Different from
the previous GAN based face completion, BoostGAN exploits both pixel level and feature level information as the supervisory
signal to preserve the identity information from multiple scales. Therefore, BoostGAN can be guaranteed to cope
with open-set occlusive profile face recognition. A new coarse-to-fine aggregation architecture with a deep GAN network (coarse net) and a shallow boosting network (fine net) is designed for identity-invariable photo-realistic face synthesis.

The main contributions of our work lie in three folds:

-We propose an end-to-end BoostGAN model for occlusive but profile face recognition, which can
synthesize photo-realistic and identity preserved frontal faces under arbitrary
poses and occlusions.

-A coarse-to-fine aggregation structure is proposed. The \textit{coarse} part is a deep GAN network for de-occlusion and frontalization of multiple partially occluded profile faces (e.g. keypoints). The \textit{fine} part is a shallow boosting network for photo-realistic face synthesis.

-We demonstrate our proposed BoostGAN model through quantitative and qualitative experiments on benchmark datasets, and
state-of-the-art results are achieved under both non-occlusive and occlusive scenarios.

\section{Related Work}
In this section, we review three closely-related topics, including generative
adversarial network (GAN), face frontalization and image completion.

\subsection{Generative Adversarial Network (GAN)}
Generative adversarial network, proposed by Goodfellow \emph{et al}.~\cite{Goodfellow2014Generative},
was formulated by gaming between a generator
and a discriminator. Deep convolutional generative adversarial network (DCGAN)~\cite{Radford2015Unsupervised}
was the first to combine CNN and GAN together for image generation. After that, many GAN variants have been proposed. However, the stability of GAN training is always an open problem, which attracts a number of researchers to solve the problem of training instability. For example, Wasserstein GAN~\cite{Arjovsky2017Wasserstein} removed the logarithm in the loss function of original GAN. Spectral normalization generative adversarial network (SNGAN)~\cite{Miyato2018Spectral} proposed the spectral normalization to satisfy the Lipschitz constant and guarantee the boundedness of statistics. Besides that, many other works focus on
how to improve the visual realism. For example, Zhu \textit{et al.} proposed the cycleGAN~\cite{Zhu2017Unpaired}
to deal with the unpaired data. Karras \textit{et al}. obtained the high-resolution image
from a low-resolution image by growing both the generator and discriminator progressively~\cite{Karras2017Progressive}.

\begin{table}
\begin{center}
 \caption{Feasibility of GAN variants for \textit{profile} face frontalization and recognition with$/$without occlusion.}
  \label{tab1}
 \small
\begin{tabular}{|c|c|c|}
\hline
Method                                       &Without occlusion                                 &With occlusion\\
\hline
DR-GAN~\cite{Luan2017Disentangled}           &\ding{51}                                   &\ding{55}\\
FF-GAN~\cite{Yin2017Towards}                 &\ding{51}                                   &\ding{55}\\
\multirow{1}{*}{TP-GAN~\cite
{Huang2017Beyond}}                           &\multirow{1}{*}{\ding{51}}                   &\ding{55} \\
\multirow{1}{*}{CAPG-GAN~\cite{Hu2018pose}}  &\multirow{1}{*}{\ding{51}}                  &\ding{55}\\

\hline
BoostGAN                                     &\ding{51}                                    &\ding{51}\\
\hline
\end{tabular}
\end{center}
\end{table}

\subsection{Face Frontalization}
Face Frontalization is an extremely challenging task due to its ill-posed nature. Existing methods dealing with this problem can be divided into three categories:
2D/3D local texture warping~\cite{Hassner2014Effective,Zhu2015High}, statistical methods
~\cite{Sagonas2015Robust}, and deep learning methods~\cite{Luan2017Disentangled, Cao2018Load,
Huang2017Beyond, Hu2018pose}. Specifically, Hassner \textit{et al}. exploited a mean 3D face reference surface
to generate a frontal view facial image for any subject~\cite{Hassner2014Effective}. Sagonas \textit{et al}. viewed the
frontal view reconstruction and landmark localization as a constrained low-rank minimization problem~\cite{Sagonas2015Robust}. Benefiting from GAN, Luan \textit{et al}. proposed a DR-GAN
for pose-invariant face recognition~\cite{Luan2017Disentangled}. FF-GAN introduced the 3D face
model into GAN, such that the 3DMM conditioned model can retain the visual quality during frontal view synthesis~\cite{Yin2017Towards}.
TP-GAN proposed to deal with profile view faces through global and local networks separately, then generate the final frontal face by fusion to improve the image photo-realistic~\cite{Huang2017Beyond}. CAPG-GAN
recovers both neutral and profile head pose face images from input face and facial landmark heatmap by pose-guided generator and couple-agent
discriminator~\cite{Hu2018pose}.

These methods work well under non-occlusive scenarios. However, they focus on the impact of pose variation, and the effect of occlusion is ignored. Due to the specificity of these network architecture, the existing methods cannot be generalized with occlusion as described in Table~\ref{tab1}.

\subsection{Image Completion}
Synthesizing the missing part of a facial image can be formulated as an image completion problem.
Content prior is required to obtain a faithful reconstruction, which usually comes from either other parts of the same image or an external dataset. The early
algorithms inpaint missing content by propagating information
from known neighborhoods based on low-level cues or global statistics,
find similar structures from the context of the input image and then paste
them in the holes~\cite{Bertalmio2003Simultaneous, Levin2003Learning,Wexler2007Space}. Deep neural network based methods repair the missing
part of the images by learning the background texture~\cite{Fawzi2016Image}. Recently, GANs have been
introduced for this task. Li \textit{et al}. designed a GAN model with global and local discriminators
for image completion~\cite{Li2017Generative}. Yeh \textit{et al}. generates the
missing content by conditioning on the available data for semantic image inpainting~\cite{Yeh2016Semantic}. However, none of them is capable of preserving the identity. Therefore, Zhao \textit{et al}. proposed to recover the missing content under various head poses while preserving the identity by
introducing an identity loss and a pose discriminator~\cite{Zhao2018Identity}.

These image completion approaches can only fill the missing region. Recognizing faces under the scenarios of both occlusion and pose variation is not well studied.

\section{Approach}
Different from those face frontalization and image completion methods, our proposed BoostGAN works on the scenario that occlusion and pose variation occur simultaneously, with regard to occlusive but profile face recognition.

\subsection{Network Architecture}
\subsubsection{Multi-occlusion Frontal View Generator}
In this work, two kinds of occlusions with regard to the position are considered: keypoint position occlusion and random position occlusion. There are four occlusive profile images $I^{b_{i}}, i\in\{1, 2, 3, 4\}$, which are covered in
left eye, right eye, nose and mouth position on a profile image by a white square
mask, respectively, for keypoint position occlusion. The corresponding frontal facial groundtruth of profile image is
denoted as $I^{gt}$. The size of $I^{b_{i}}$, $i\in\{1, 2, 3, 4\}$ and $I^{gt}$ are
$W\times H\times C$, where $W$, $H$, and $C$ are defined as width, height, and channel of image,
respectively. The aim of the multi-occlusion frontal view generator is to recover four rough but slightly
discriminative frontal images from the four different occlusive profile images. The multi-occlusion
frontal view generator $G^{c}$ is composed of an encoder and decoder, denoted as $\{G^{c}_{e}$,
$G^{c}_{d}\}$. That is,
\begin{equation}
I^{f_{i}}=G^{c}(I^{b_{i}}), i\in\{1, 2, 3, 4\},
\label{eq1}
\end{equation}
where $I^{f_{i}}, i\in\{1, 2, 3, 4\}$ is the generated frontal image with respect to each different occlusive profile image, respectively.

Inspired by the excellent performance of TP-GAN~\cite{Huang2017Beyond}, our $G^{c}$ follows the
similar network structure, that is formulated with a down-sampling encoder and an up-sampling
decoder with skip connections, for fair comparison and model analysis.
\begin{table}
\begin{center}
 \caption{Configuration of the boosting network.}
  \label{tab2}
 \small
\begin{tabular}{|c|c|c|c|}
\hline
 Layer & Input & Filter Size & Output Size \\
\hline
\hline
\multirow{2}{*}{resblock1} & concatenated& \multirow{2}{*}{$\begin{bmatrix} 5\times5,12 \\ 5\times5,12 \end{bmatrix}\times1$ }& \multirow{2}{*}{$128\times128\times12 $ }\\
& image & & \\
\hline
conv1 & resblock1 & $5\times5, 64$ & $128\times128\times64$ \\
\hline
resblock2 & conv1 & $\begin{bmatrix} 3\times3,64 \\ 3\times3,64 \end{bmatrix}\times1$ & $128\times128\times64 $ \\
\hline
conv2 & resblock2 & $3\times3, 32$ & $128\times128\times32$ \\
\hline
conv3 & conv2 & $3\times3, 3$ & $128\times128\times3$ \\
\hline
\end{tabular}
\end{center}
\end{table}

\subsubsection{Multi-input Boosting Network}
Through the multi-occlusion frontal view generator $G^{c}$, four rough and slight discriminative
frontal faces can be coarsely obtained. Due to the occlusion of key regions, the effectiveness of identity
preservation is not good and the photo-realistic of synthesis is flawed with blur and distortion. Therefore, a multi-input boosting network is then followed for photo-realistic and identity preservation, because we find that the four inputs are complementary. Boosting is an ensemble
meta-algorithm, for transforming a family of weak learners into a strong one by fusion. Therefore, the four primarily generated facial images by the generator can be further converted to a photo-realistic and identity preserving frontal face by the boosting network.

The boosting network denoted as $G^{f}$ is used to
deal with image generation. Simply, the four primary generated facial images are concatenated as
the input (the size is $W\times H\times 4C$) of the boosting network. That is,
\begin{equation}
I^{f}=G^{f}(I^{f_{1}}, I^{f_{2}}, I^{f_{3}}, I^{f_{4}}),
\label{eq2}
\end{equation}
where $I^{f}$ denotes the final generated photo-realistic and identity preserving frontal face.
The boosting network contains two residual blocks, with each follows a convolutional layer. In order
to avoid overfitting and vanishing gradient, Leaky-ReLU and batch normalization are used. The detail of network configuration is shown in Table~\ref{tab2}.

\subsection{Training Loss Functions}
The proposed BoostGAN is trained by taking a weighted sum of different losses as the supervisory signal in an end-to-end manner,
including adversarial loss, identity preserving loss, and pixel-level losses.

\subsubsection{Adversarial Loss}
In the training stage, two components: discriminator $D$ and generator $G$ are included, where the goal of $D$ is to distinguish the fake data produced by
generator $G^{c}$ and $G^{f}$ from the real data. The generator $G^{c}$ and $G^{f}$ aim
to synthesize realistic-looking images to fool $D$. The game between $G$ and $D$ can
be represented as a value function $V(D, G)$:
\begin{equation}
\begin{split}
\min_{G^{c},G^{f}}\max_{D}V(D,G)=&\frac{1}{N}\sum^{N}_{n=1}\{\textrm{log}D(I^{gt}_{n})+ \frac{1}{5}(\sum^{4}_{i=1}\textrm{log}(1- \\
&D(G^{c}(I^{b_{i}}_{n})))+\textrm{log}(1-D(G^{f}(I^{f_{1}}_{n},\\
&I^{f_{2}}_{n},I^{f_{3}}_{n},I^{f_{4}}_{n}))))\},
\end{split}
\end{equation}
where $N$ is the batch size. In practice, $G^{c}$, $G^{f}$ and $D$ are
alternatively optimized via the following objectives:
\begin{equation}
\begin{split}
\max_{D}V(D,G)=&\frac{1}{N}\sum^{N}_{n=1}\{\textrm{log}D(I^{gt}_{n})+ \frac{1}{5}(\sum^{4}_{i=1}\textrm{log}(1- \\
&D(G^{c}(I^{b_{i}}_{n})))+\textrm{log}(1-D(G^{f}(I^{f_{1}}_{n},\\
&I^{f_{2}}_{n},I^{f_{3}}_{n},I^{f_{4}}_{n}))))\}.
\end{split}
\end{equation}

\begin{equation}
\begin{split}
L_{adv}=\max_{G^{c},G^{f}}V(D,G)=&\frac{1}{5N}\sum^{N}_{n=1}\{\sum^{4}_{i=1}\textrm{log}D(G^{c}(I^{b_{i}}_{n}))+\\
&\textrm{log}D(G^{f}(I^{f_{1}}_{n},I^{f_{2}}_{n},I^{f_{3}}_{n},I^{f_{4}}_{n})))\}.
\end{split}
\end{equation}

\subsubsection{Identity Preserving Loss}
During the de-occlusion and frontalization process, facial identity information is easy to be lost, which is undoubtedly harmful to face recognition performance. Therefore, for better preservation of human identity of the generated image, the identity-wise feature representation with a pre-trained face recognition network is used as supervisory signal. The similarity measured with L1-distance formulates the identity preserving loss function:
\begin{equation}
L_{ip}=|F^{po}(I^{p})-F^{po}(\hat{I})|+|F^{fc}(I^{p})-F^{fc}(\hat{I})|,
\end{equation}
where $F^{po}(\cdot)$ and $F^{fc}(\cdot)$ denote the output of the last pooling layer and
the fully connected layer of the pre-trained Light CNN~\cite{Wu2015A} on large faces datasets, respectively.
$\hat{I}$ represents all generated faces in BoostGAN,
including $I^{f_{i}}, i\in\{1, 2, 3, 4\}$, and $I^{f}$.

\subsubsection{Pixel-level Losses}
In order to guarantee the multi-image content consistency and improve the photo-realistic, three pixel-level losses,
i.e. multi-scale pixel-wise L1 loss, symmetry loss, and total variation regularization~\cite{Johnson2016Perceptual} are employed:
\begin{equation}
L_{pix}=\frac{1}{K}\sum^{K}_{k=1}\frac{1}{W_{k}H_{k}C}\sum_{w,h,c=1}^{W_{k},H_{k},C}|\hat{I}_{s,w,h,c}-I_{s,w,h,c}^{gt}|,
\end{equation}
\begin{equation}
L_{sym}=\frac{1}{W/2\times H}\sum^{W/2}_{w=1}\sum^{H}_{h=1}|\hat{I}_{w,h}-\hat{I}_{w_{s},h}|,
\end{equation}
\begin{equation}
L_{tv}= \sum^{C}_{c=1}\sum^{W,H}_{w,h=1}|\hat{I}_{w+1,h,c}-\hat{I}_{w,h,c}|+ |\hat{I}_{w,h+1,c}-\hat{I}_{w,h,c}|,
\end{equation}
where $K$ denotes the number of scales. $W_{k}$ and $H_{k}$ denote the width and height of
each image scale, respectively. $w_{s}=W-(w-1)$ is the symmetric
abscissa of $w$ in $\hat{I}$. In our proposed approach, three
scales $(32\times32, 64\times64$, and $128\times128)$ are considered.

\subsubsection{Overall Loss of Generator}
In summary, the ultimate loss function for training the generator $G$ ($G^{c}$ and $G^{f}$) is a weighted summation
of the above loss functions:
\begin{equation}
L_{syn} = \lambda_{1}L_{adv}+\lambda_{2}L_{pix}+\lambda_{3}L_{sym}+\lambda_{4}L_{ip}+\lambda_{5}L_{tv},
\end{equation}
where $\lambda_{1},\lambda_{2},\lambda_{3},\lambda_{4}$ and $\lambda_{5}$ are trade-off
parameters.

\section{Experiments}
To demonstrate the effectiveness of our proposed method for occlusive profile face recognition,
the qualitative and quantitative experiments on benchmark datasets (constrained vs. unconstrained) are conducted in this
paper. For the former,
we show the qualitative results of frontal face synthesis under various poses and occlusions.
For the latter, face recognition performance on the occlusive profile faces across different poses and occlusions is evaluated.

\begin{figure}
\centering
\subfigure[Profile]{
\includegraphics[width=1.4cm]{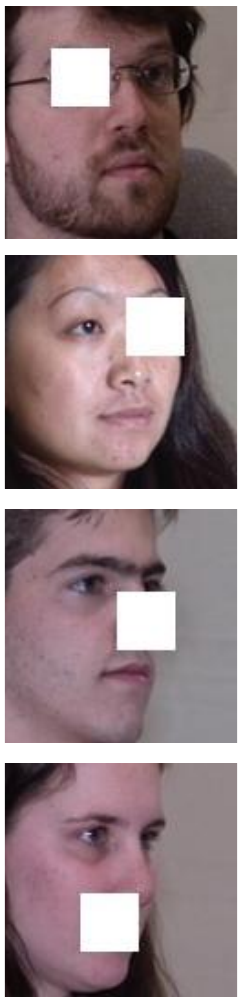}}
\subfigure[\textbf{Ours}]{
\includegraphics[width=1.4cm]{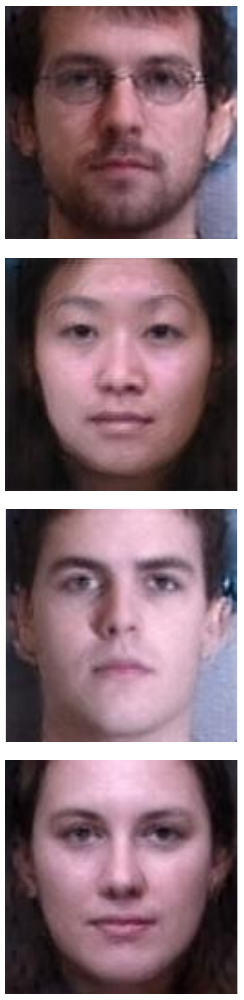}}
\subfigure[\cite{Luan2017Disentangled}]{
\includegraphics[width=1.4cm]{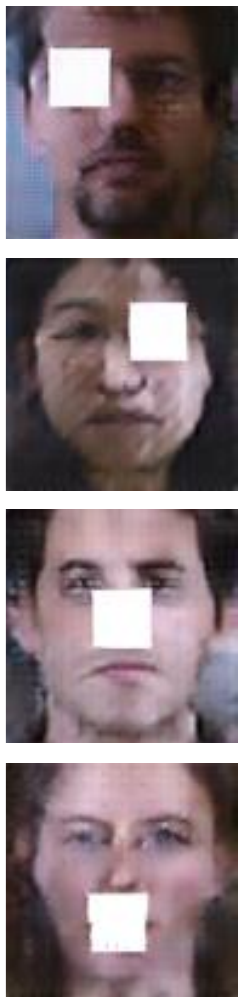}}
\subfigure[\cite{Huang2017Beyond}*]{
\includegraphics[width=1.414cm]{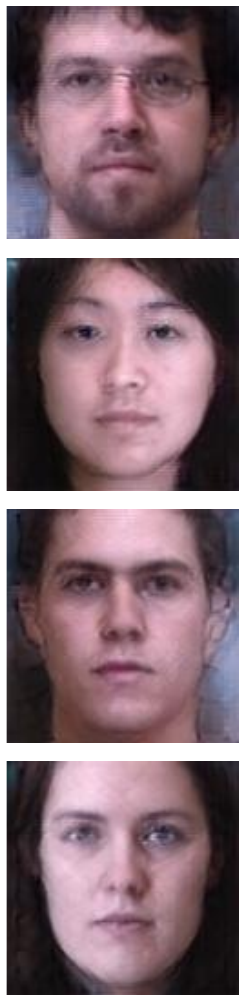}}
\subfigure[GT]{
\includegraphics[width=1.4cm]{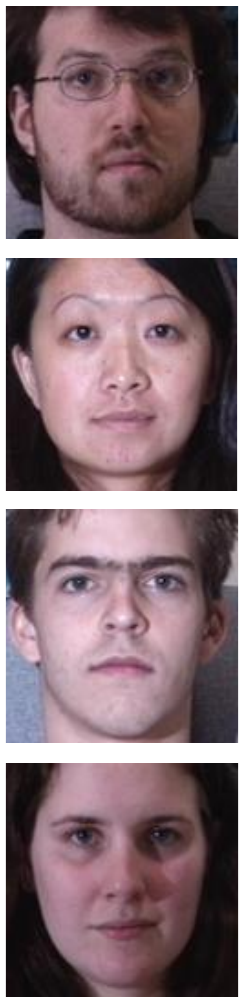}}
\caption{Synthesis results on \textbf{{\color{red}keypoint region}} occluded Multi-PIE dataset. From top to bottom, the poses are $15^{\circ}$,
$30^{\circ}$, $45^{\circ}$, $60^{\circ}$. The ground truth frontal images are provided at the last column.}
\label{fig4}
\end{figure}

\begin{figure}
\centering
\subfigure[Profile]{
\includegraphics[width=1.4cm]{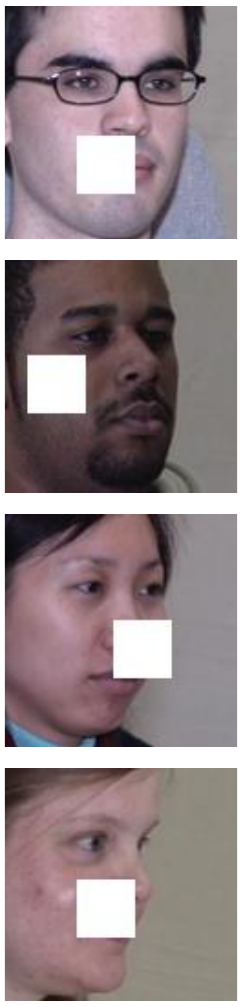}}
\subfigure[\textbf{Ours}]{
\includegraphics[width=1.4cm]{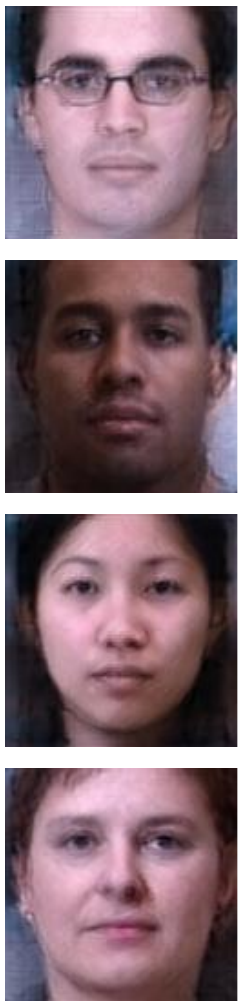}}
\subfigure[\cite{Luan2017Disentangled}]{
\includegraphics[width=1.4cm]{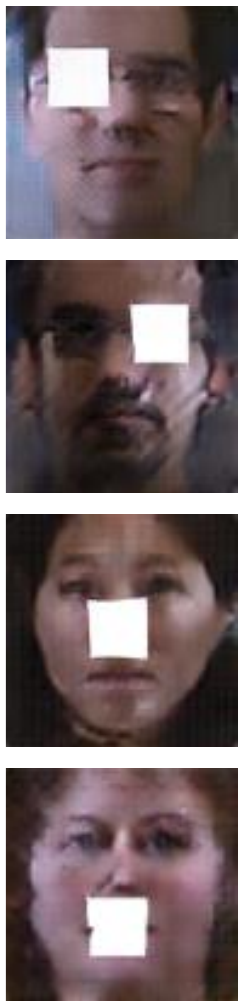}}
\subfigure[\cite{Huang2017Beyond}*]{
\includegraphics[width=1.414cm]{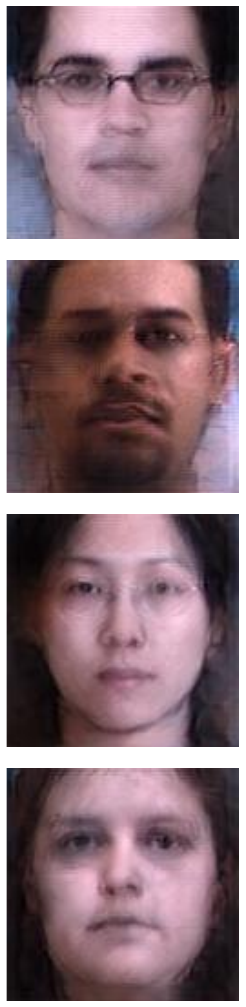}}
\subfigure[GT]{
\includegraphics[width=1.4cm]{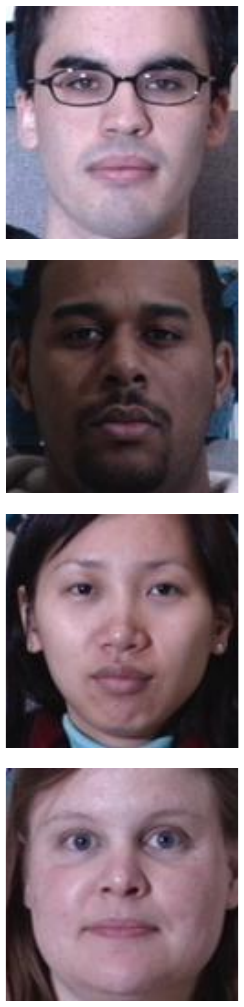}}
\caption{Synthesis results on \textbf{{\color{red}random block}} occluded Multi-PIE dataset. From top to bottom, the poses are $15^{\circ}$,
$30^{\circ}$, $45^{\circ}$, $60^{\circ}$. The ground truth frontal images are provided at the last column. Note that the all the models are trained solely on keypoint occluded Multi-PIE dataset.}
\label{fig5}
\end{figure}

\subsection{Experimental Settings}
\textbf{Databases.} Multi-PIE~\cite{Gross2010Multi} is the largest database for evaluating face
recognition and synthesis in a constrained setting. A total of 337 subjects were recorded in
four sessions. 20 illumination levels and 13 poses range from $-90^{\circ}$
to $90^{\circ}$ are included per subject. Following the testing protocol in~\cite{Luan2017Disentangled,Huang2017Beyond},
we use 337 subjects with neutral expression and 11 poses within $\pm75^{\circ}$ from all
sessions. The first 200 subjects are used as training set and the remaining 137 subjects are used
as testing set. In testing stage, the first appearance image with frontal and neutral illumination per subject is viewed as gallery, and the others are probes.

The LFW~\cite{Huang2007Labled} database contains 13,233 images from 5,749 subjects, in which only
85 subjects have more than 15 images, and 4069 people have only one image. It is generally used to
evaluate the face verification or synthesis performance in the wild (i.e. unconstrained setting). Following the face
verification protocol~\cite{Huang2007Labled}, the 10-fold cross validation strategy is
considered for verification performance evaluation on the generated images. Several state-of-the-art models such as FF-GAN~\cite{Yin2017Towards}, DR-GAN~\cite{Luan2017Disentangled}, and TP-GAN~\cite{Huang2017Beyond} have been compared
with our approach.

\textbf{Data Preprocessing.} In order to guarantee the generality of BoostGAN and reduce the model parameter bias, both Multi-PIE and LFW are detected by MTCNN~\cite{Zhang2016Joint} and aligned
to a canonical view of size $128\times128$. Two kinds of occlusions: keypoint occlusion and random occlusion across positions are used in our
work. For the former, the centers of occlusion masks are the
facial key points, i.e. left eye, right eye, tip of the nose, and the center of mouth. For the latter, the centers of occlusion masks are randomly positioned.
The size of each occlusion is $32\times32$ (the keypoint region can be completely covered) filled with white pixels.

\textbf{Implementation Details.} In conventional GAN~\cite{Goodfellow2014Generative}, Goodfellow
\textit{et al}. suggested to alternate between $k$ (usually $k=1$) steps of optimizing $D$ and
one step of optimizing $G$. Thus, we update 2 steps for optimizing $G^{s}$ and $G^{f}$, and 1 for
$D$, ensuring a good performance. In all experiments, we set $\lambda_{1}=
2e1,\lambda_{2}=1,\lambda_{3}=3e-1$, $\lambda_{4}=4e1$ and $\lambda_{5}=1e-3$.

\begin{figure}
\centering
\subfigure[Profile]{
\includegraphics[width=1.4cm]{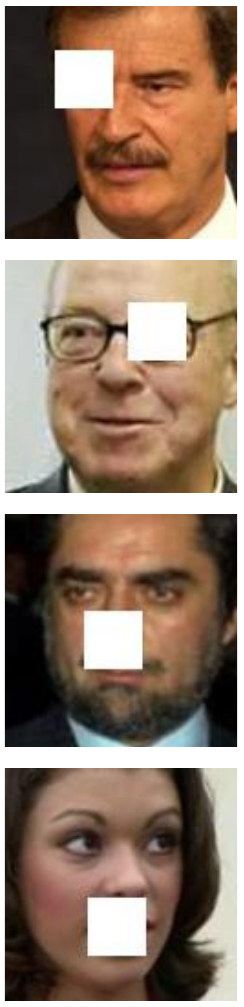}}
\subfigure[\textbf{Ours}]{
\includegraphics[width=1.4cm]{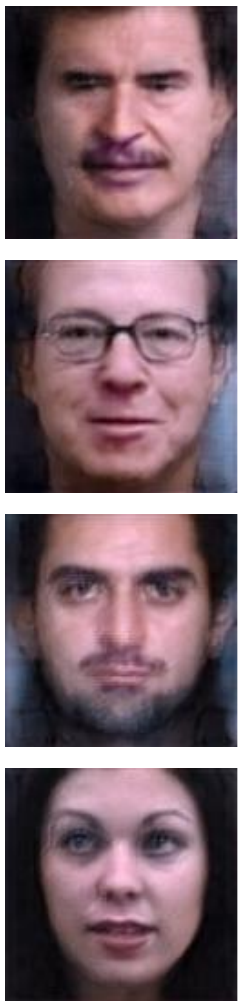}}
\subfigure[\cite{Luan2017Disentangled}]{
\includegraphics[width=1.4cm]{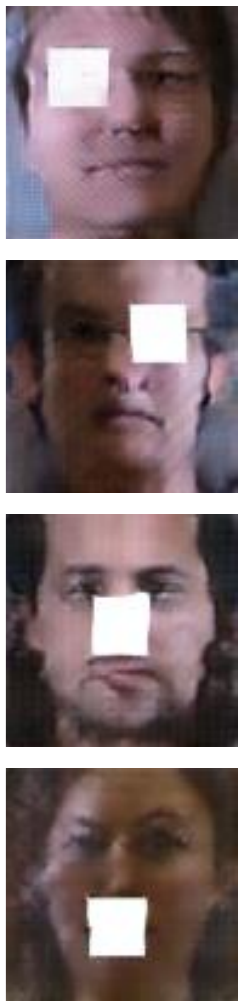}}
\subfigure[\cite{Huang2017Beyond}*]{
\includegraphics[width=1.414cm]{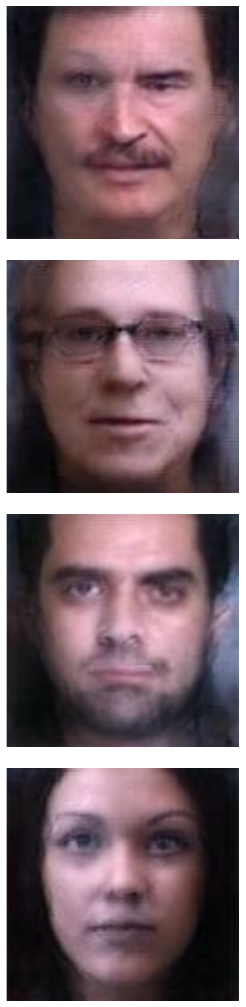}}
\caption{Synthesis results on \textbf{{\color{red}keypoint region}} occluded LFW dataset in the wild. Note that there are no ground truth frontal images for this dataset. The models are solely trained based on keypoint occluded Multi-PIE dataset.}
\label{fig6}
\end{figure}

\subsection{Face Frontalization}
To qualitatively demonstrate the synthetic ability of our method, the generative frontal images under
different poses and occlusions in different settings are shown in this section. The qualitative
experiments of face synthesis are divided into 3 parts: occlusive Multi-PIE, occlusive LFW, and Multi-PIE after de-occlusion.

\begin{figure}
\centering
\subfigure[Profile]{
\includegraphics[width=1.4cm]{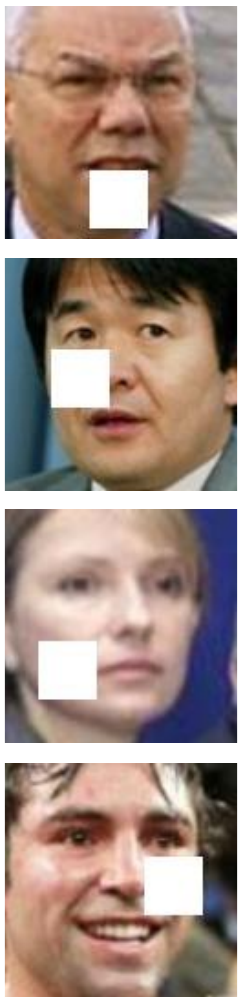}}
\subfigure[\textbf{Ours}]{
\includegraphics[width=1.4cm]{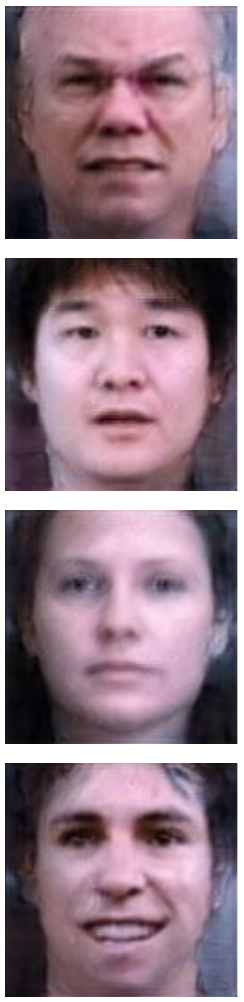}}
\subfigure[\cite{Luan2017Disentangled}]{
\includegraphics[width=1.4cm]{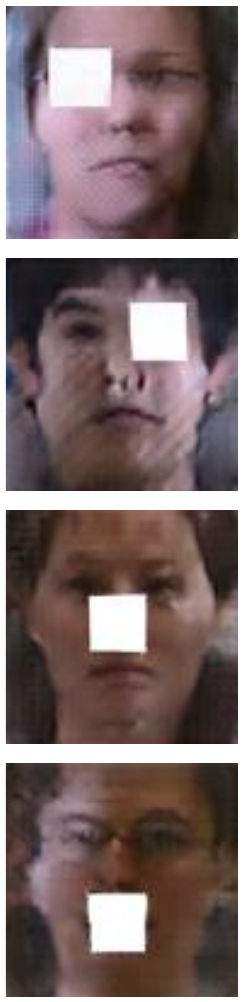}}
\subfigure[\cite{Huang2017Beyond}*]{
\includegraphics[width=1.414cm]{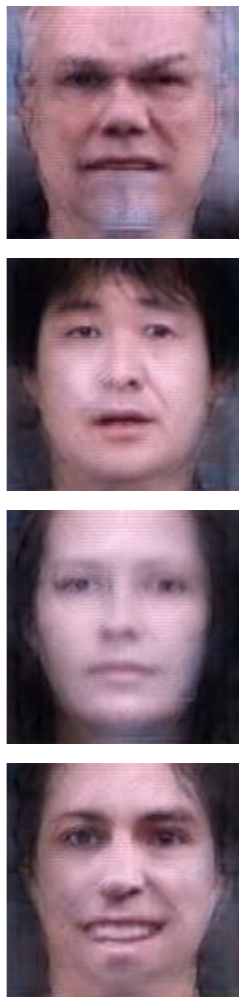}}
\caption{Synthesis results on \textbf{{\color{red}random block}} occluded LFW dataset. Note that all the models are trained solely on keypoint occluded Multi-PIE dataset, without retraining on randomly blocked datasets.}
\label{fig7}
\end{figure}

\begin{figure*}
\centering
\subfigure[Profile]{
\includegraphics[width=1.49cm]{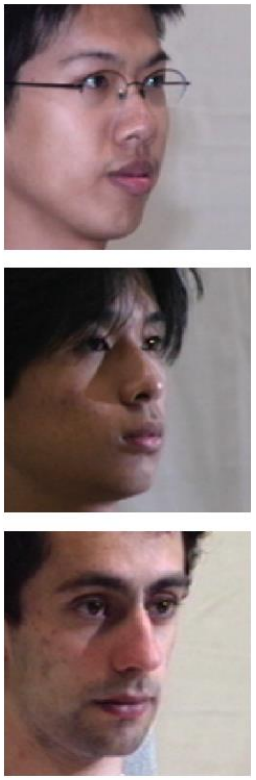}}
\subfigure[\textbf{Ours}]{
\includegraphics[width=1.495cm]{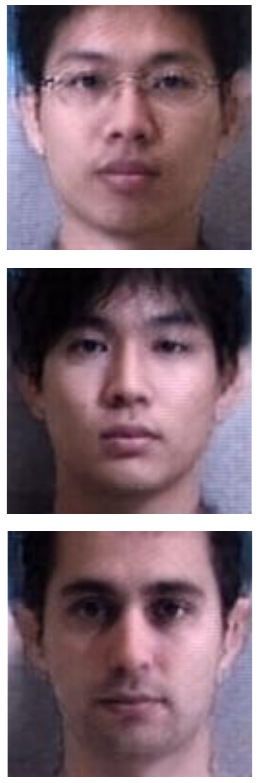}}
\subfigure[\cite{Huang2017Beyond}]{
\includegraphics[width=1.5cm]{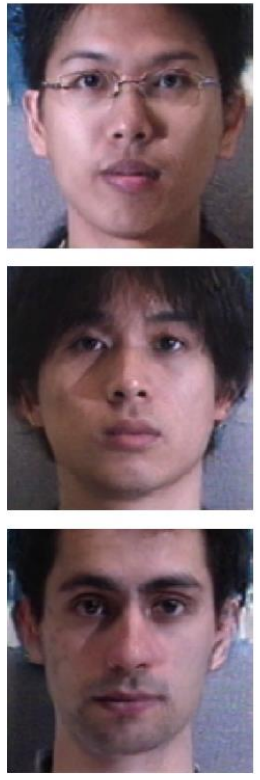}}
\subfigure[\cite{Luan2017Disentangled}]{
\includegraphics[width=1.5cm]{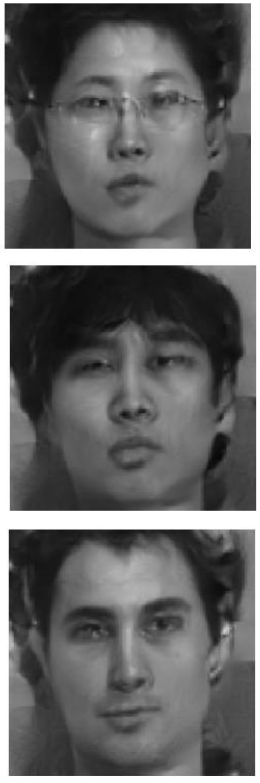}}
\subfigure[\cite{Yim2015Rotating}]{
\includegraphics[width=1.48cm]{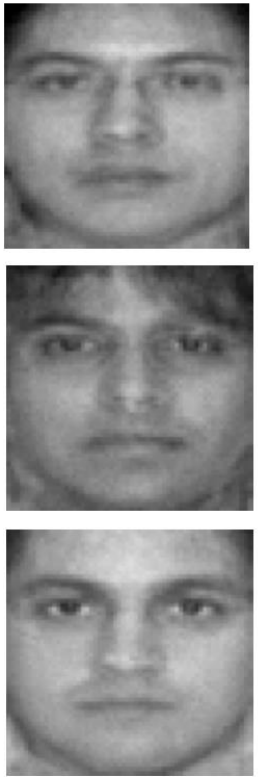}}
\subfigure[\cite{Ghodrati2015Towards}]{
\includegraphics[width=1.48cm]{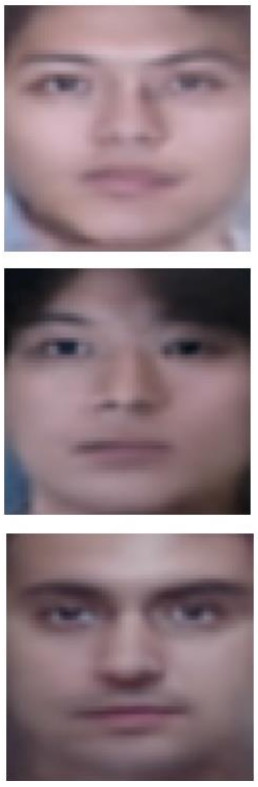}}
\subfigure[\cite{Zhu2015High}]{
\includegraphics[width=1.475cm]{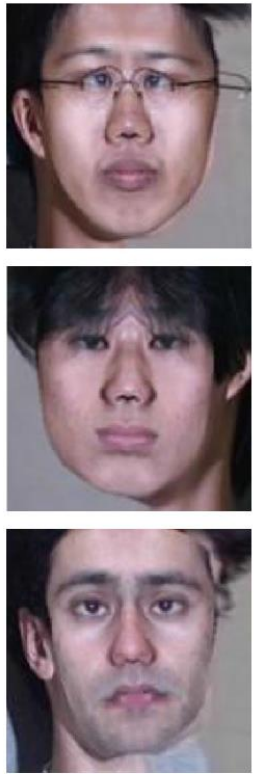}}
\subfigure[\cite{Hassner2014Effective}]{
\includegraphics[width=1.5cm]{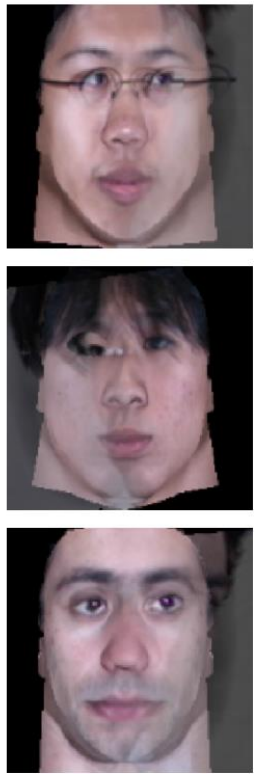}}
\subfigure[GT]{
\includegraphics[width=1.495cm]{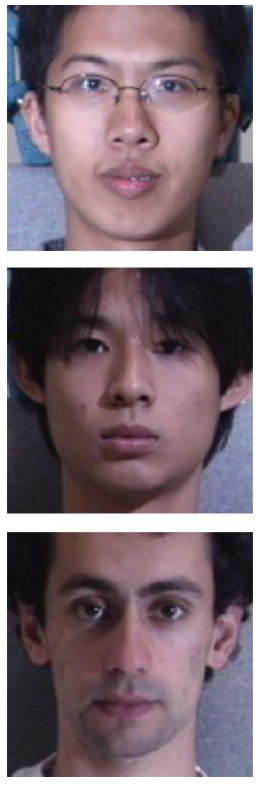}}
\caption{Comparison with state-of-the-art synthesis methods under the pose variation of $45^{\circ}$ (the first two rows) and $30^{\circ}$
(the last row). The BoostGAN model is trained on non-occlusive Multi-PIE dataset and the generality for non-occlusive frontalization is demonstrated.}
\label{fig3}
\end{figure*}

\textbf{Face Synthesis on Occlusive Multi-PIE.} With the two different types of block occlusions, the synthesis results are shown in
Figure~\ref{fig4} and Figure~\ref{fig5}, respectively. Note that for each block occlusion the trained model with keypoint occlusion is used. We can see that the proposed model can generate photo-realistic and identity preserving faces under occlusions, which are better than DR-GAN and TP-GAN. Due to the missing of keypoint region pixels, the landmarks located patch network of TP-GAN~\cite{Huang2017Beyond}
become unavailable, and only the global parametric model can be trained on occlusive Multi-PIE. For convenience, we mark it as TP-GAN*. Specially, due to the lack of supervision of ground truth frontal image in DR-GAN model,
it can not fill the hole (block occlusion) in the facial image.

Notably, the synthesis results of random occlusive profile image are shown in Figure~\ref{fig5}. Obviously,
the generated images of BoostGAN are still better than DR-GAN and TP-GAN*. The synthesis
performance of both DR-GAN and TP-GAN is degraded due to the random occlusive.
The generated results of TP-GAN* become more distorted due to that the random occlusive profile images may not appeared in training process. Additionally, with random occlusion, the position of occlusion in the generated images by DR-GAN is also changed as some keypoint position appeared in the training process.
However, different from DR-GAN and TP-GAN*, the proposed BoostGAN can still obtain good synthetic images.
Further, with the increasing of pose angle, BoostGAN can faithfully synthesize frontal view images
with clear and clean details.

\textbf{Face Synthesis on Occlusive LFW.} In order to demonstrate the generalization ability in
the wild, the LFW database is used to test BoostGAN model trained solely on the keypoint region occlusive
Multi-PIE database. As shown in Figure~\ref{fig6} and Figure~\ref{fig7}, BoostGAN also obtains
better visual results on LFW than others, but the background color is similar to Multi-PIE. This is understandable because the model is trained solely on Multi-PIE.

\textbf{Face Synthesis on Multi-PIE after De-occlusion.} We compare the synthesis results of BoostGAN against
state-of-art face synthesis methods on faces after de-occlusion in Figure~\ref{fig3}. The de-occlusive Multi-PIE here denotes
the original non-occlusive Multi-PIE. It can be seen that, the generated results of the proposed BoostGAN still show the effectiveness even without occlusions. Note that BoostGAN shows competitive performance with TP-GAN~\cite{Huang2017Beyond}, but obviously better than other methods
no matter in global structure or local texture. This demonstrates that although BoostGAN is proposed under the occlusion scenarios, it can also works well under non-occlusive scenario and the generality is further verified.

\subsection{Face Recognition}

\begin{table}
\begin{center}
 \caption{Rank-1 recognition rate (\%) comparison on \textit{keypoint region} occluded Multi-PIE. \textbf{Black}: ranks the $1^{st}$; \textbf{\color{red}Red}: ranks the $2^{nd}$; \textbf{\color{blue}Blue}: ranks the $3^{rd}$.}
  \label{tab4}
\small
\begin{tabular}{c c c c c c}
\hline
Method                             & $\pm15^{\circ}$ & $\pm30^{\circ}$ & $\pm45^{\circ}$ & $\pm60^{\circ}$ & $\pm75^{\circ}$\\
\hline
\hline
DR-GAN~\cite{Luan2017Disentangled}
(k1)                               &  67.38          &  60.68          & 55.83           & 47.25           &  39.34       \\
DR-GAN~\cite{Luan2017Disentangled}
(k2)                               &  73.24          &  65.37          & 59.90           & 51.18           &  \textbf{{\color{red}42.24}}       \\
DR-GAN~\cite{Luan2017Disentangled}
(k3)                               &  66.93          &  60.60          & 56.54           & 49.70           &  39.77       \\
DR-GAN~\cite{Luan2017Disentangled}
(k4)                               &  71.33          &  63.72          & 57.59           & 50.10           &  \textbf{{\color{blue}40.87}}       \\
DR-GAN~\cite{Luan2017Disentangled}
(mean)                             &  69.72          &  62.59          & 57.47           & 49.56           &  40.55       \\
\hline
TP-GAN~\cite{Huang2017Beyond}*
(k1)                               &  \textbf{{\color{blue}98.17}}          &  \textbf{{\color{blue}95.46}}          & \textbf{{\color{blue}86.60}}           & \textbf{{\color{blue}65.91}}           &  39.51       \\
TP-GAN~\cite{Huang2017Beyond}*
(k2)                               &  \textbf{{\color{red}99.27}}          &  \textbf{{\color{red}97.25}}          & \textbf{{\color{red}88.37}}           & \textbf{{\color{red}66.03}}           &  40.82       \\
TP-GAN~\cite{Huang2017Beyond}*
(k3)                               &  95.04          &  90.95          & 82.72           & 62.40           &  38.67       \\
TP-GAN~\cite{Huang2017Beyond}*
(k4)                               &  97.80          &  93.66          & 83.84           & 62.27           &  36.76       \\
TP-GAN~\cite{Huang2017Beyond}*
(mean)                             &  97.57          &  94.33          & 85.38           & 64.15           &  38.94       \\
\hline
BoostGAN                           &  \textbf{99.48} &  \textbf{97.75} & \textbf{91.55}  & \textbf{72.76}  &  \textbf{48.44}\\                  \end{tabular}
\end{center}
\end{table}

\begin{table}
\begin{center}
 \caption{Rank-1 recognition rate (\%) comparison on \textit{random block} occluded Multi-PIE. \textbf{Black}: ranks the $1^{st}$; \textbf{\color{red}Red}: ranks the $2^{nd}$; \textbf{\color{blue}Blue}: ranks the $3^{rd}$.}
  \label{tab5}
\small
\begin{tabular}{c c c c c c}
\hline
Method                             & $\pm15^{\circ}$ & $\pm30^{\circ}$ & $\pm45^{\circ}$ & $\pm60^{\circ}$ & $\pm75^{\circ}$\\
\hline
\hline
DR-GAN~\cite{Luan2017Disentangled}
(r1)                               &  47.64	         &  38.93	       & 33.21	         & 25.38           &  18.92       \\
DR-GAN~\cite{Luan2017Disentangled}
(r2)                               &  65.75          &  55.15          & 46.52           & 38.33           &  \textbf{{\color{blue}29.00}}      \\
DR-GAN~\cite{Luan2017Disentangled}
(r3)                               &  56.01	         &  46.27	       & 39.13	         & 29.11	       &  23.01       \\
DR-GAN~\cite{Luan2017Disentangled}
(r4)                               &  59.10	         &  47.92	       & 39.97	         & 33.69     	   &  25.20       \\
DR-GAN~\cite{Luan2017Disentangled}
(mean)                             &  57.13	         &  47.07	       & 39.71	         & 31.63	       &  24.03       \\
\hline
TP-GAN~\cite{Huang2017Beyond}*
(r1)                               &  \textbf{{\color{red}89.81}}          &  \textbf{{\color{red}83.88}}          & \textbf{{\color{red}74.94}}           & \textbf{{\color{red}54.83}}           &  \textbf{{\color{red}31.34}}       \\
TP-GAN~\cite{Huang2017Beyond}*
(r2)                               &  77.98          &  71.68          & 60.52           & 42.68           &  23.92       \\
TP-GAN~\cite{Huang2017Beyond}*
(r3)                               &  79.12          &  72.45          & 60.00           & 41.37           &  24.11       \\
TP-GAN~\cite{Huang2017Beyond}*
(r4)                               &  \textbf{{\color{blue}86.13}}          &  \textbf{{\color{blue}77.76}}          & 64.84           & 45.08           &  25.15       \\
TP-GAN~\cite{Huang2017Beyond}*
(mean)                             &  83.26          &  76.44          & \textbf{{\color{blue}65.08}}           & \textbf{{\color{blue}45.99}}           &  26.13       \\
\hline
BoostGAN                           &  \textbf{99.45} &  \textbf{97.50} & \textbf{91.11}  & \textbf{72.12}  &  \textbf{48.53}\\                  \end{tabular}
\end{center}
\end{table}

The proposed BoostGAN aims to recognize human faces with occlusions and pose variations. Therefore, for verifying the identity preserving capacity of different models, face recognition on benchmark datasets is studied. We first use the trained generative models to frontalize the profile
face images in Multi-PIE and LFW, then evaluate the performance of face recognition or verification
by using the Light CNN extracted features of those generated frontal facial images.
Similar to the qualitative experiments, the quantitative experiments include 3 parts: face
recognition on occlusive Multi-PIE, face verification on occlusive LFW, and face recognition on Multi-PIE after de-occlusion.

\textbf{Face Recognition on Occlusive Multi-PIE.} Similarly, the trained model is solely on the keypoint region occluded images. The rank-1 recognition rates for the two types of occlusions are shown in Table~\ref{tab4} and Table~\ref{tab5}, respectively. $\textrm{k}i, i\in{1,2,3,4}$ denotes
the four different block mask regions, such as left eye, right eye, nose and mouth. $\textrm{r}
i, i\in{1,2,3,4}$ denotes four random block occlusions.

It is obvious that the performance of BoostGAN outperforms DR-GAN and TP-GAN* for each type of blocked occlusion. It is common that with the increase of pose angles, the recognition performance is decreased. However, compared with other methods, BoostGAN still shows state-of-the-art performance. Specially, by comparing Table~\ref{tab5} with
Table~\ref{tab4}, we observe that the recognition rates of DR-GAN and TP-GAN* show a dramatically decrease due to changes of occlusive types. However, the proposed BoostGAN is almost unaffected.

\begin{table}
\begin{center}
 \caption{Face verification accuracy (ACC) and area-under-curve (AUC) results on \textit{keypoint region} occluded LFW.}
  \label{tab6}
\small
\begin{tabular}{c c c}
\hline
Method                             & ACC(\%)         & AUC(\%) \\
\hline
\hline
DR-GAN~\cite{Luan2017Disentangled}
(k1)                               &  67.60          &   73.65  \\
DR-GAN~\cite{Luan2017Disentangled}
(k2)                               &  67.28	         &   72.94  \\
DR-GAN~\cite{Luan2017Disentangled}
(k3)                               &  58.43	         &   59.19  \\
DR-GAN~\cite{Luan2017Disentangled}
(k4)                               &  69.50	         &   76.05  \\
DR-GAN~\cite{Luan2017Disentangled}
(mean)                             &  65.71	         &   70.46  \\
\hline
TP-GAN~\cite{Huang2017Beyond}*
(k1)                               &  86.52          &   92.81  \\
TP-GAN~\cite{Huang2017Beyond}*
(k2)                               &  \textbf{{\color{red}87.83}}          &   \textbf{{\color{blue}93.96}}  \\
TP-GAN~\cite{Huang2017Beyond}*
(k3)                               &  85.17          &   91.63  \\
TP-GAN~\cite{Huang2017Beyond}*
(k4)                               &  \textbf{{\color{blue}87.78}}          &   \textbf{{\color{red}93.97}}  \\
TP-GAN~\cite{Huang2017Beyond}
(mean)                             &  86.83          &   93.09  \\
\hline
BoostGAN                           &  \textbf{89.57} &   \textbf{94.90}  \\
\end{tabular}
\end{center}
\end{table}

\begin{table}
\begin{center}
 \caption{Face verification accuracy (ACC) and area-under-curve (AUC) results on \textit{random block} occluded LFW.}
  \label{tab7}
\small
\begin{tabular}{c c c}
\hline
Method                             & ACC(\%)         & AUC(\%) \\
\hline
\hline
DR-GAN~\cite{Luan2017Disentangled}
(r1)                               &  63.28          &   67.20  \\
DR-GAN~\cite{Luan2017Disentangled}
(r2)                               &  65.53	         &   71.79  \\
DR-GAN~\cite{Luan2017Disentangled}
(r3)                               &  57.15	         &   57.76  \\
DR-GAN~\cite{Luan2017Disentangled}
(r4)                               &  64.82	         &   70.35  \\
DR-GAN~\cite{Luan2017Disentangled}
(mean)                             &  62.70	         &   66.78  \\
\hline
TP-GAN~\cite{Huang2017Beyond}*
(r1)                               &  \textbf{{\color{blue}82.75}}          &   \textbf{{\color{blue}89.86}}  \\
TP-GAN~\cite{Huang2017Beyond}*
(r2)                               &  77.65          &   84.63  \\
TP-GAN~\cite{Huang2017Beyond}*
(r3)                               &  81.07          &   88.24  \\
TP-GAN~\cite{Huang2017Beyond}*
(r4)                               &  \textbf{{\color{red}83.25}}          &   \textbf{{\color{red}90.15}}  \\
TP-GAN~\cite{Huang2017Beyond}*
(mean)                             &  81.18          &   88.22  \\
\hline
BoostGAN                           &  \textbf{89.58} &   \textbf{94.75}  \\
\end{tabular}
\end{center}
\end{table}

\textbf{Face Verification on Occlusive LFW.} Face verification performance evaluated on the recognition accuracy (ACC) and area under ROC curves (AUC) in the wild are provided in Table~\ref{tab6} and Table~\ref{tab7}, based on two types of occlusions. From the results, we observe that DR-GAN is seriously flawed due to the model's weak specificity to occlusions. TP-GAN has shown comparable results in Table~\ref{tab6} under keypoint occlusion, but significantly degraded performance in Table~\ref{tab7} under random occlusion. However, similar to constrained Multi-PIE dataset, the proposed BoostGAN shows state-of-the-art performance under occlusions, and there is no performance degradation across different types of occlusion. We can conclude that our BoostGAN shows excellent generalization power for occlusive but profile face recognition in the wild.

\textbf{Face Recognition on Multi-PIE after De-occlusion.} After discussion of the recognition performance under occlusions, we further verify the effectiveness of the proposed method under profile but non-occlusive faces. The Rank-1 accuracies of different methods on Multi-PIE
database are presented in Table~\ref{tab3}. Specifically, 8 methods including FIP+LDA~\cite{Zhu2013Deep}, MVP+LDA~\cite
{Zhu2014Multi}, CPF~\cite{Yim2015Rotating}, DR-GAN~\cite{Luan2017Disentangled}, $\textrm{DR-GAN}_{\textrm{AM}}$~\cite{Luan2018Rep}, FF-GAN~\cite{Yin2017Towards} and TP-GAN~\cite
{Huang2017Beyond} are compared. The results of Light CNN are used as the baseline. All the methods are following the same experimental protocol for fair comparison. We can observe that the proposed BoostGAN outperforms all other methods for clean profile face recognition. TP-GAN~\cite{Huang2017Beyond}, as the state-of-the-art method, is also inferior to ours.

\textbf{Discussion.} Our approach is an ensemble model to complete the facial de-occlusion and frontalization
simultaneously, aiming at face recognition tasks under large pose variations and occlusions. Although BoostGAN achieves a success in the uninvestigated synthesis problem under occlusion, some similar characteristics with existing GAN variants are equipped.
First, the encoder-decoder based CNN architecture is used. Second, the traditional loss functions in pixel level and feature level are
exploited. Third, the basic components such as generator and discriminator of GAN are used. The key difference between ours and other
GAN variants lie in that the complementary information out of occlusion is boosted. Due to the space limitation, more experiments on different sized occlusions are explored in \textbf{Supplementary Material}.
\begin{table}
\begin{center}
 \caption{Rank-1 recognition rate (\%) comparison on profile Multi-PIE without occlusion.}
  \label{tab3}
\small
\begin{tabular}{c c c c c c}
\hline
Method                             & $\pm15^{\circ}$ & $\pm30^{\circ}$ &  $\pm45^{\circ}$ &  $\pm60^{\circ}$ &  mean \\
\hline
\hline
FIP+LDA~\cite{Zhu2013Deep}         &  90.7           &  80.7           &  64.1            &  45.9            &  70.35  \\
MVP+LDA~\cite{Zhu2014Multi}        &  92.8           &  83.7           &  72.9            &  60.1            &  77.38  \\
CPF~\cite{Yim2015Rotating}         &  95.0           &  88.5           &  79.9            &  61.9            &  81.33  \\
DR-GAN~\cite{Luan2017Disentangled} &  94.0           &  90.1           &  86.2            &  83.2            &  88.38   \\
$\textrm{DR-GAN}_{\textrm{AM}}$~\cite{Luan2018Rep}
                                   &  95.0           &  91.3           &  88.0            &  \textbf{{\color{blue}85.8}}            &  90.03   \\
FF-GAN~\cite{Yin2017Towards}       &  94.6           &  92.5           &  89.7            &  85.2            &  \textbf{{\color{blue}90.50}}  \\
TP-GAN~\cite{Huang2017Beyond}      &  \textbf{{\color{red}98.68}}          &  \textbf{{\color{red}98.06}}          &  \textbf{{\color{red}95.38}}           &  \textbf{87.72}  &  \textbf{{\color{red}94.96}}  \\
\hline
Light CNN~\cite{Wu2015A}           &  \textbf{{\color{blue}98.59}}          &  \textbf{{\color{blue}97.38}}          &  \textbf{{\color{blue}92.13}}           &  62.09           &  87.55  \\
BoostGAN                           &  \textbf{99.88} &  \textbf{99.19} &  \textbf{96.84}  &  \textbf{{\color{red}87.52}}  & \textbf{95.86}\\
\end{tabular}
\end{center}
\end{table}

\section{Conclusion}
This paper has answered \textit{how to recognize faces if large pose variation and occlusion exist simultaneously}. Specifically, we contribute a BoostGAN model for occlusive but profile face recognition in constrained and unconstrained settings. The proposed model follows an end-to-end training protocol, from a multi-occlusion frontal view generator to a multi-input boosting network, and achieves coarse-to-fine de-occlusion and frontalization. The adversarial generator aims to realize coarse frontalization, de-occlusion and identity preservation across large pose variations and occlusions. The boosting network targets at generating photo-realistic, clean and frontal faces by ensemble the complementary information of multiple inputs. Extensive experiments on benchmark datasets have shown the generality and superiority of the proposed BoostGAN over other state-of-the-art under occlusive and non-occlusive scenarios.

{\small
\bibliographystyle{ieee}
\bibliography{egbib}
}

\end{document}